\documentclass{llncs}
\usepackage[margin=1.5in,footskip=0.25in]{geometry}
 \usepackage[colorlinks = true,
            linkcolor = blue,
            urlcolor  = blue,
            citecolor = red,
            anchorcolor = blue]{hyperref}
            
\usepackage{graphicx}
\usepackage{multirow}
\usepackage{amsmath}
\usepackage{pdflscape}
\usepackage{rotating}
\usepackage[compress]{cite}

\begin{document}

\title{Deep learning for word-level handwritten \\Indic script identification}
\titlerunning{Deep learning for Indic script identification}
\author{
Soumya Ukil\inst{1} \and 
Swarnendu Ghosh\inst{1} \and
Sk Md Obaidullah\inst{2} \and 
K. C. Santosh\inst{3} \and 
Kaushik Roy\inst{4} \and 
Nibaran Das\inst{1} 
}
\authorrunning{S. Ukil, S. Ghosh, S.M. Obaidullah, K. C. Santosh, K. Roy, N. Das}

\institute{
Dept. of Computer Science \& Engineering, Jadavpur University, Kolkata 700032, WB, India \and
Dept. of Computer Science \& Engineering,  Aliah University, Kolkata 700156, WB, India \and 
Dept. of Computer Science, The University of South Dakota, Vermillion, SD 57069, USA \and 
Dept. of Computer Science \& Engineering,  West Bengal State University, 700126, WB, India \\
\mbox{}\\
Corresponding authors: K. C. Santosh (\email{santosh.kc@usd.edu})  \& N. Das (\email{nibaran@gmail.com}) \hfill \mbox{}\\
}
\maketitle

\begin{abstract}
We propose a novel method that uses {\em c}onvolutional {\em n}eural {\em n}etworks (CNNs) for feature extraction. Not just limited to conventional spatial domain representation, we use multilevel 2D discrete Haar wavelet transform, where image representations are scaled to a variety of different sizes. These are then used to train different CNNs to select features. To be precise, we use 10 different CNNs that select a set of 10240 features, i.e. 1024/CNN. With this, 11 different handwritten scripts are  identified, where 1K words per script are used. In our test, we have achieved the maximum script identification rate of 94.73\% using {\em m}ulti-{\em l}ayer {\em p}erceptron (MLP). Our results outperform the state-of-the-art techniques.

\end{abstract}

\section{Introduction}
\begin{figure}[tbp]
\includegraphics[width = 0.5\textwidth]{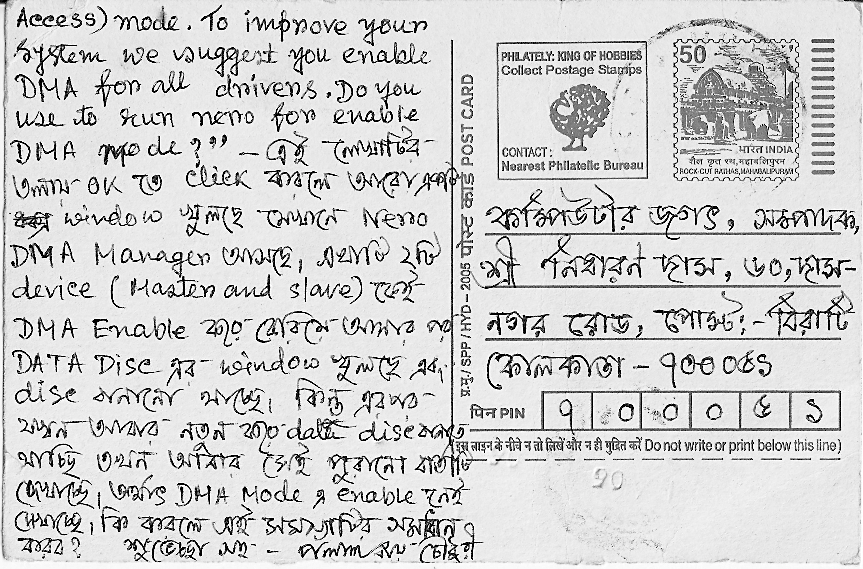} \includegraphics[width = 0.5\textwidth]{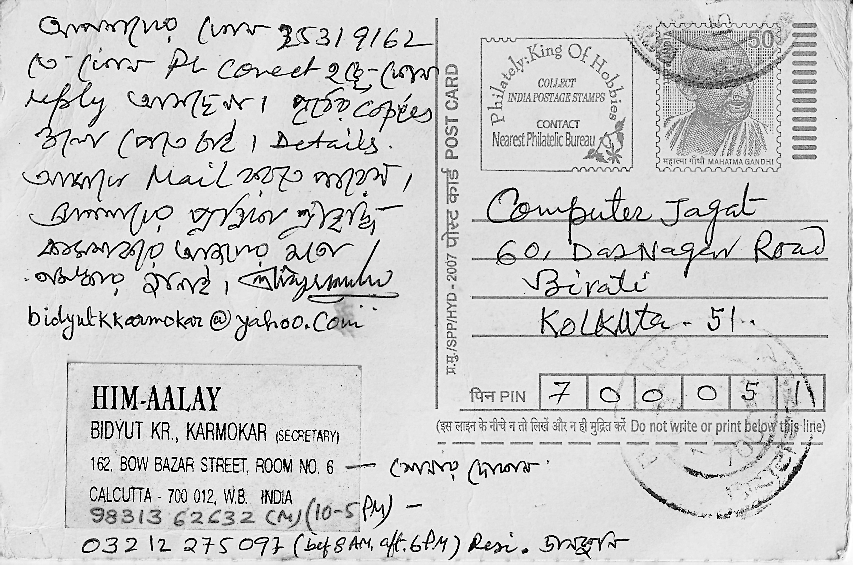}
\caption{Two multi-script postal document images, where Bangla, Roman and Devanagari scripts are used.}
\label{fig:1}
\end{figure}

{\em O}ptical {\em c}haracter {\em r}ecognition (OCR) has always been a challenging field in pattern recognition. OCR techniques are used to convert handwritten or machine printed scanned document images to machine-encoded texts. These OCR techniques are script dependent. Therefore, script identification is considered as a precursor to OCR. In particular, in case of a multi-lingual country like India script identification is the must since a single document, such as postal documents and business forms, contains several different scripts (see Fig.~\ref{fig:1}).

Indic handwritten script identification has a rich state-of-the-art literature~\cite{ghosh2010script,pal2012handwriting,singh2015word,hangarge2013directional}. More often, previous works have been focusing on word-level script identification~\cite{PATI20081218}. Not stopping there, in a recent work~\cite{obaidullah2017phdindic_11}, authors introduced page-level script identification performance to see whether we can expedite the processing time. In general, in their works, hand-crafted features that are based on structural and/or visual appearances (morphology-based) were used. The question is, are we just relying on what we see and use apply features accordingly or can we just let machine to select features that are required for optimal identification rate? This inspires to use deep learning, where CNNs can be used for extracting and/or selecting features for identification task(s).

Needless to say, CNNs have stood well with their immense contribution in the field of OCR. Their onset has ben marked by the ground-breaking performance of CNNs on MNIST dataset~\cite{lecun1998gradient}. Very recently, the use CNN for Indic script (Bangla character recognition) has been reported ~\cite{roy2017handwritten}. Not to be confused, the primary of goal of this paper is to use deep learning concept to identify 11 different handwritten Indic scripts: Bangla, Devnagari, Gujarati, Gurumukhi, Kannada, Malayalam, Oriya, Roman, Tamil, Telugu and Urdu. Inspired from deep learning-based concept, we use CNNs to select features from handwritten document images (scanned), where we use multilevel 2D discrete Haar wavelet transform (in addition to conventional spatial domain representation) and image representations are scaled to a variety of different sizes. With these representation, several different CNNs are used to select features. In short, the primary idea behind this is to avoid using hand-crafted features for identification. Using {\em m}ulti-{\em l}ayer {\em p}erceptron (MLP), 11 different handwritten scripts (as mentioned earlier) are identified with satisfactory performance.

The remainder of the paper can be summarized as follows. Section~\ref{sec:pa} provides a quick overview of our contribution, where it includes CNN architecture and feature extraction process. In Section~\ref{sec:exp}, experimental results are provided. It also includes a quick comparison study. Section~\ref{sec:conc} concludes our paper.

\section{Contribution outline}\label{sec:pa}
As mentioned earlier, in stead of using hand-crafted features for document image representation, our goal is to let deep learning to select distinguishing features for optimal script identification. For a few but recent works, where CNNs have used with successful classification, we refer to~\cite{lecun1998gradient,krizhevsky2012imagenet,sarkhel2017multi}. We observe that CNNs work especially when we have sufficient data to train. This means, data redundancies will be helpful. In general, CNN takes raw pixel data (image) and as training proceeds, the model learns distinguishing features that can successfully contribute to identification/classification. Such a training process produces a feature vector that summarize the important aspects of the studied image(s). 

More precisely, our approach is twofold: first, we use a two- and three-layered CNNs for three different scales of the input image; and secondly, we use exactly same CNNs for two different scales of the transformed image (wavelet transform). We then merge those features and make ready for script identification. In what follows, we explain our CNN architecture including definitions, parameters for wavelet transform and the way we produce features.

\begin{figure}[tbp]
\includegraphics[width = \textwidth]{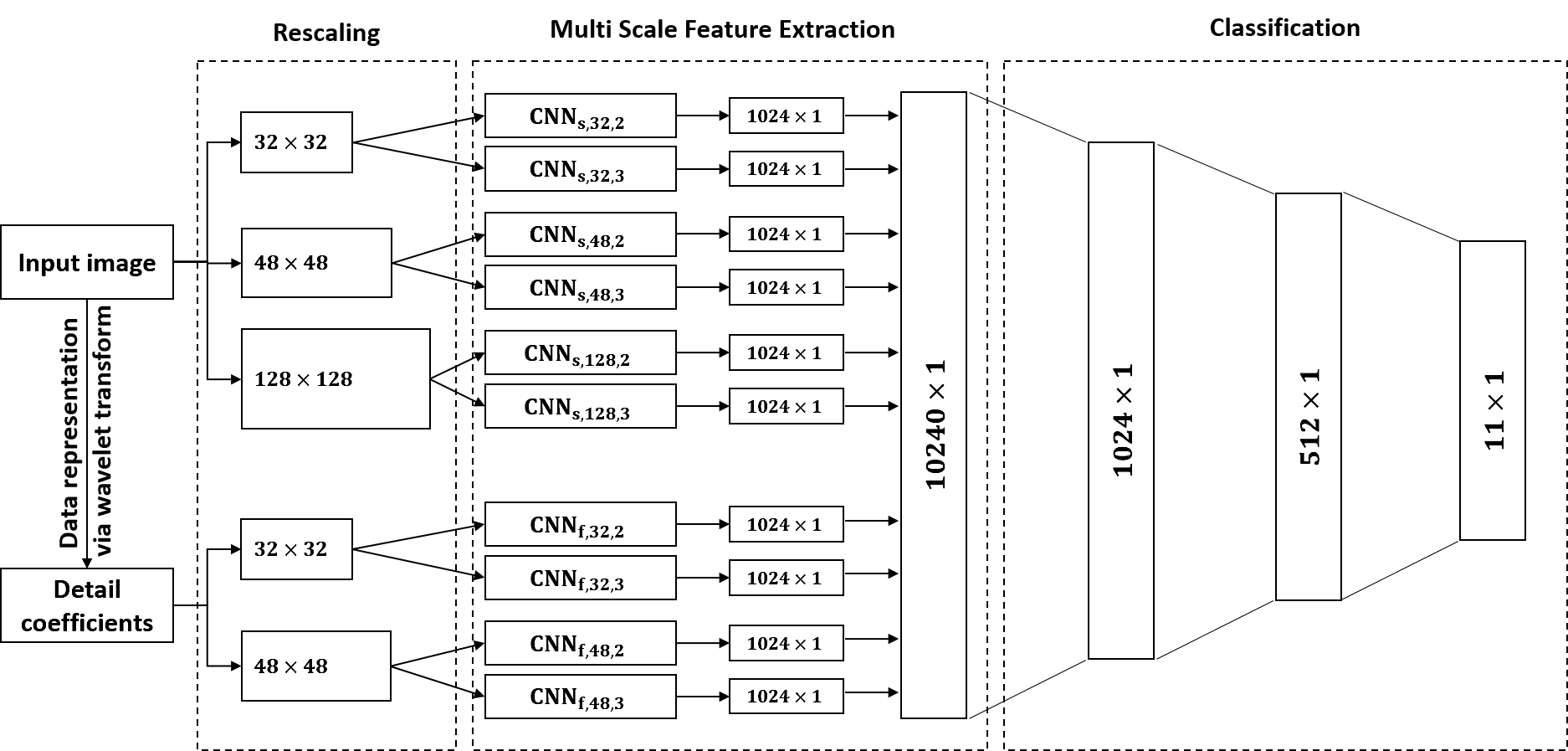}
\caption{Schematic block diagram of handwritten Indic script identification showing different modules: feature extraction/selection and classification.}
\label{fig:flowchart}
\end{figure}

\subsection{CNN architecture}\label{sec:cnn}
In general, a CNN has a layered architecture consisting of three basic types of layers namely, \begin{enumerate}
\item {\em c}onvolutional {\em l}ayer (CL), 
\item {\em p}ooling {\em l}ayer (PL) and 
\item {\em f}ully {\em c}onnected {\em l}ayer (FCL).
\end{enumerate}
CLs consist of a set of kernels that produce parameters and help in convolution operation. In CL, every kernel generates an activation map as an output. PLs do not have parameters but, their major role is to avoid possible data redundancies (still preserving their significance). In our approach, all CNNs have max-pooling operation at their corresponding PLs. In addition to these two different types of layers, FCL is used, where MLP has been in place. 

In Fig.~\ref{fig:flowchart}, we provide a complete schematic block diagram of handwritten Indic script identification showing different modules: feature extraction/selection and classification. In our study, 10 different CNNs are used to select features from a variety of representations of the studied image and we label each of them as $\text{CNN}_{d,x,y}$. In every $\text{CNN}_{d,x,y}$, $d$ and $x$ respectively refer to domain representation and dimension of the studied image, and $y$ refers to the number of convolutional and pooling layers in that particular CNN. For example, domain representation can be expressed as $d = \{s, f \}$, where $s$ refers to spatial domain and $f$, frequency. In our case, either of these is taken into account. Note that, with the use of the {\em H}aar {\em w}avelet {\em t}ransform (HWT) (see Section~\ref{sec:wt}), certain frequencies are removed. In case of dimension ($x$), we have $x = \{[32\times32], [48 \times 48], [128 \times 128] \}$, and for simplicity, $x = \{ 32, 48, 128\}$ is used. All of these dimensions signify resolution to which the input images are scaled. In case of CL and PL, $y=\{2,3 \}$: one of the two is taken in CNN, and $y=2$ means that there are two pairs of convolutional and pooling layers in the CNNs. In our model, two broad CNN architectures: $\text{CNN}_{d,x,2}$ and $\text{CNN}_{d,x,3}$ are used and can be summarized as in Table~\ref{tab:arch}.

\begin{enumerate}
\item $\text{CNN}_{d,x,2}$: We have six different layers, i.e. two CLs, two PLs and two FCLs. The first two CLs are followed by PLs. In FCLs, the first layer takes image representation that has been generated by CLs and PLs and reshapes them in the form of a vector, and the second FCL produces a set of 1024 features. 

\item $\text{CNN}_{d,x,3}$: Every CNN has eight different layers: three CLs, three PLs and two FCLs. In general, such an architecture is very much similar to previously mentioned $\text{CNN}_{d,x,2}$. The difference lies in additional pair of CL and PL that follows second pair of the same. Like the $\text{CNN}_{d,x,2}$, these CNNs produce a set of 1024 features for any studied image. 
\end{enumerate}

\begin{sidewaysfigure}
\centering
\includegraphics[width = \textwidth]{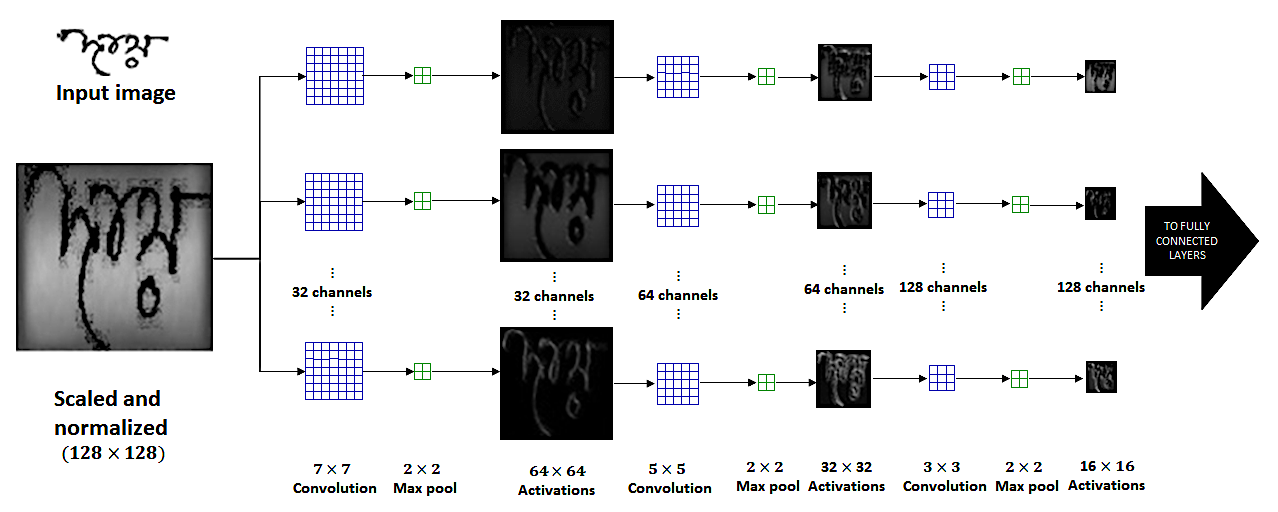}
\caption{Illustrating the activation maps for $\text{CNN}_{s,128,3}$: spatial domain image representation with the dimensionality of 128 and three-layered convolutional and pooling layers.}
\label{fig:activations}
\end{sidewaysfigure}

Once again, the architectural details of aforementioned CNNs are summarized in Table~\ref{tab:arch} that follows schematic block diagram of the system (see Fig.~\ref{fig:flowchart}). For more understanding, in Fig.~\ref{fig:activations}, we provide the activation maps for $\text{CNN}_{s,128,3}$. This means that it uses spatial domain image representation with the dimensionality of 128 and three pairs of convolutional and pooling layers in the CNNs.

\begin{table}[tbp]
\centering
\renewcommand{\tabcolsep}{0.5em}
\renewcommand{\arraystretch}{1.2}
\caption{Architecture: $\text{CNN}_{d,x,y}$, where $y=\{2,3 \}$.}
\label{tab:arch}
\begin{tabular}{llccccccccc}
\hline
&& \multicolumn{7}{c}{Layer}\\ 
Architecture & Parameter  & CL1 &  PL1 & CL2 &  PL2 & CL3 &  PL3 &FCL1 &  FCL2 & Softmax\\ 
\hline \hline
$\text{CNN}_{d,x,2}$
& Channel 	& 32 & 32 & 64 & 64 &--- &--- & 1024 & 512 & 11\\ 
& Filter size & 5$\times$5 & 2$\times$2 & 5$\times$5 & 2$\times$2 & --- & ---  & ---  & ---  & ---\\
& Pad size   & 2  & --- & 2 & --- & --- &--- &--- &--- &---\\
\hline
$\text{CNN}_{d,x,3}$
& Channel 	& 32 & 32 & 64 & 64 & 128 & 128 & 1024 & 512 & 11\\ 
& Filter size & 7$\times$7 & 2$\times$2 & 5$\times$5 & 2$\times$2 & 3$\times$3 & 2$\times$2& --- & ---  & --- \\
& Pad size   & 3  & --- & 2 & --- & 1 &--- &--- &--- &--- \\ \hline
\multicolumn{9}{l}{\underline{Index}}\\ 
\multicolumn{9}{l}{CL = {\em c}onvolutional {\em l}ayer}\\
\multicolumn{9}{l}{PL = {\em p}ooling {\em l}ayer}\\
\multicolumn{9}{l}{FCL = {\em f}ully {\em c}onnected {\em l}ayer}\\
\end{tabular}
\end{table}

\subsection{Data representation}\label{sec:wt}
In general, since Fourier transform~\cite{smith1997scientist} may not work to successfully provide information about what frequencies are present at what time, {\em w}avelet {\em t}ransform (WT) ~\cite{daubechies1990wavelet} and {\em s}hort {\em t}ime {\em F}ourier {\em t}ransform~\cite{portnoff1980time} are typically used. Both of these help identify the frequency components present in any signal at any given time. WT, on the other hand, can provide dynamic resolution.

We consider an image a 2D time signal that can be resized.  We then use multilevel 2D discrete WT on a scaled/resized image ($128 \times 128$) to generate frequency domain representation. To be precise, we use the {\em H}aar {\em w}avelet~\cite{sundararajan2011fundamentals} with seven different level decomposition that can generate approximated and detailed coefficients. Since the approximated coefficients are equivalent to zero, we use the detailed coefficients in addition to modified approximated coefficients to reconstruct the image. In the modified approximated coefficients, we consider only high frequency components. 

In our method, using a variety of different WTs, such as the Daubechies~\cite{vonesch2007generalized} and several decomposition levels, the best results were observed with the Haar wavelet and a decomposition level of 7. This reconstructed image is further resized to $32 \times 32$ (x = 32) and $48 \times 48$ (x = 48), and are fed into multiple CNNs as mentioned in Section~\ref{sec:cnn}. Like in the frequency domain representation, spatial domain representations are resized/scaled to $32 \times 32$ (x = 32), $48 \times 48$ (x=48) and $128 \times 128$ (x =128) are fed into multiple CNNs.

\section{Experiments}\label{sec:exp}
\subsection{Dataset, and evaluation metrics and protocol.}\label{sec:data}
\begin{figure}[tbp]
\begin{center}
\includegraphics[width = 0.9\textwidth]{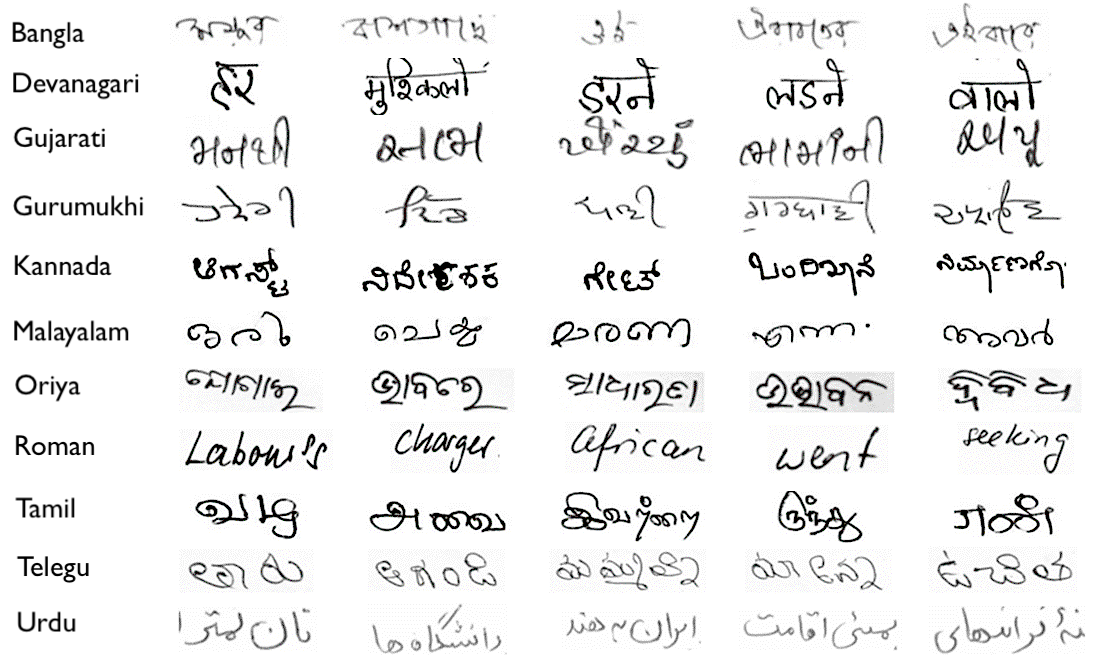}
\caption{Illustrating few samples from dataset named PHDIndic\_11 used in our experiment.}
\label{fig:data}
\end{center}
\end{figure}

To evaluate our proposed approach for word-level handwritten script identification, we have considered a dataset named PHD\_Indic\_11~\cite{obaidullah2017phdindic_11}. It is composed of 11K scanned word images (grayscale) from 11 different Indic script, i.e. 1K per script. A few samples are shown in Fig.~\ref{fig:data}. For more information about dataset, we refer to recently reported work~\cite{obaidullah2017phdindic_11}. The primary reason behind considering PHD\_Indic\_11 dataset in our test is, no big size data has been reported in the literature for research purpose, till this date.

Using the exact same CNN representation as before, $\text{C}_{d,x,y}[i][j]$ represents the count where an instance with label $i$ is classified as $j$. Accuracy (acc) of the particular CNN can then be computed as
\begin{flalign}\nonumber
&\text{acc}_{d,x,y} = \frac{\sum_{i=1}^{11}{\text{C}_{d,x,y}[i][i]}}{\sum_{i=1}^{11}{\sum_{j=1}^{11}{\text{C}_{d,x,y}[j][i]}}}.&
\end{flalign}
Precision (prec) can be computed as
\begin{flalign}\nonumber
&\text{prec}_{d,x,y} = \frac{\sum_{i=1}^{11}{\text{prec}_{d,x,y} ^{i}}}{11} \text{ and } \text{prec}_{d,x,y}^{i}  = \frac{ \text{C}_{d,x,y}[i][i] } { \sum_{j=1}^{11} \text{C}_{d,x,y}[i][j] }, &
\end{flalign}
where $\text{prec}_{d,x,y}^{i}$ refers to precision for any $i$-th label. In a similar fashion, recall (rec) can be computed as
\begin{flalign}\nonumber
&\text{rec}_{d,x,y} = \frac{\sum_{i=1}^{11}{\text{rec}_{d,x,y} ^{i}}}{11} \text{ and } \text{rec}_{d,x,y}^{i}  = \frac{ \text{C}_{d,x,y}[i][i] } { \sum_{j=1}^{11} \text{C}_{d,x,y}[i][j] }, &
\end{flalign}
where $\text{rec}_{d,x,y}^{i}$ refers to recall for any $i$-th label. Having both precision and recall, f-score can be computed as 
\begin{flalign}\nonumber
& \text{f-score}_{d,x,y} = \frac{\sum_{i=1}^{11}{\text{f-score}_{d,x,y}^{i}}}{11} \text{ and } \text{f-score}_{d,x,y}^{i}  = 2 \times \frac{ \text{prec}_{d,x,y}^i \times  \text{rec}_{d,x,y}^i  }  { \text{prec}_{d,x,y}^i  +  \text{rec}_{d,x,y}^i  }. &
\end{flalign}

Following conventional 4:1, i.e. train:test evaluation protocol, we have separated 8.8K images for training and the remaining 2.2K images for testing. We ran our experiments by using a machine: GTX 730 with 384 CUDA cores and 4GB GPU RAM. Besides, it has Intel Pentium Core2Quad Q6600 and 4GB RAM.

\subsection{Experimental set up}\label{sec:setup}
As mentioned earlier, are are required to train the CNNs first before testing.  In other words, it is important to see how training and testing have been performed. 

Our CNNs in this study represented by $\text{CNN}_{d,x,y}$ are trained independently using the training dataset (as mentioned earlier). To clarify once again, these CNNs has either two or three pairs of consecutive convolutional and pooling layers. Besides, each of them has three fully connected layers. The first of these three layers function as an input layer with number of neurons that depends on the size of the input image specified by $x$. The second layer in CNNs has 1024 neurons, and during training, we apply a dropout probability of 0.5. The final layer has 11 neurons, whose outputs are used as input to an 11-way soft-max classifier that provides us with classification/identification probabilities for each of 11 possible classes. Our training set of 8.8K word images are split into multiple batches of 50 word images and the CNNs are trained accordingly. For optimization, Adam optimizer~\cite{kingma2014adam} was used with learning rate of $1 \times 10 ^{-3}$ having default parameters: $\beta_1 = 0.9$  and $\beta_2   = 0.999$. This helps apply gradients for loss to the weight parameters during back propagation. We computed the accuracy of the CNN as training proceeds by taking ratio of the  of images successfully classified in the batch to the total number of images in the studied batch.

After training CNNs with 8.8K word images, we evaluated/tested each of them independently with the test set that is composed of 2.2K word images.  

More specifically, for each input size specification, we have two CNNs, i.e. for domain (d) and input size ($x \times x$): $\text{CNN}_{d,x,2}$ and $\text{CNN}_{d,x,3}$. Altogether, we have 10 different CNNs since we have three different input sizes ($x= \{32, 48, 128\}$) for raw image and two different input sizes  ($x= \{32, 48\}$) for wavelet transformed image/data. For better understanding, we refer readers to Fig.~\ref{fig:flowchart}. Note that we have trained the CNNs to extract 1024 features from each one, i.e. $120 \times 1$. These are then concatenated to form a single $10240 \times 1$ vector, where ten different CNNs are employed. Like in the conventional machine learning classification, these features are used for training and testing purpose using MLP classifier.  

\begin{figure}[tbp]
\begin{center}
\includegraphics[width = 0.95\textwidth]{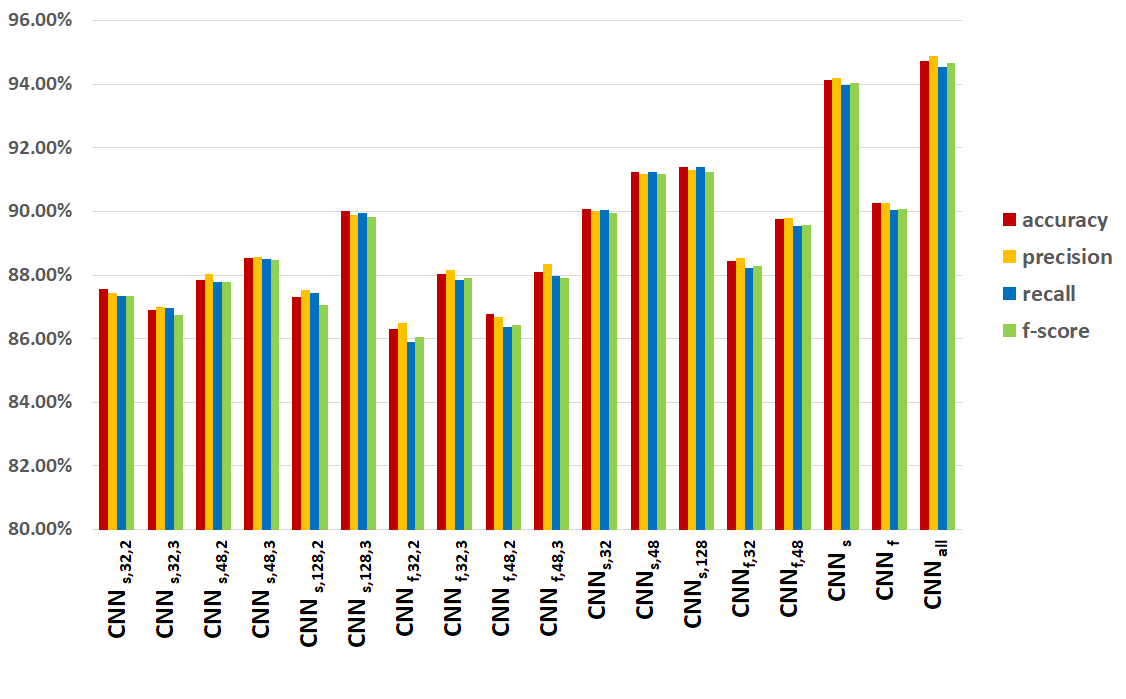}
\caption {Our results (in terms of accuracy, precision, recall and f-score) for all networks: CNNs and their possible combinations.}
\label{fig:result}
\end{center}
\end{figure}

\subsection{Our results and comparative study}\label{sec:cs}
In this section, using dataset, and evaluation metrics (see Section~\ref{sec:data}), and experimental setup (see Section~\ref{sec:setup}), we summarize our results and comparative study as follows:
\begin{enumerate}
\item We provide results that have been produced from different architectures (CNNs), and select the highest script identification rate from them;  and
\item We then take highest script identification for a comparative study, where previous relevant works are considered.
\end{enumerate}

\begin{figure}[tbp]
\begin{center}
\includegraphics[width = 0.65\textwidth]{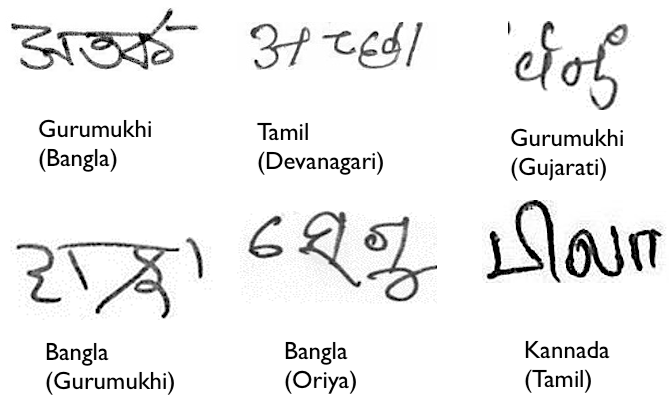}
\caption{Misclassified samples, where script names in the bracket are the actual scripts but, our system identified them incorrectly. For example, in the first case, word image has been identified as Gurumukhi  and it actually is Bangla.}
\label{fig:miss}
\end{center}
\end{figure}

\subsubsection{Our results:}
Fig.~\ref{fig:result} shows the comparison of the individual CNNs along with the effect of combining them. The individual $\text{CNN}_{d,x,y}$ produced the maximum script identification rate of 90\% for $\text{CNN}_{s,128,3}$. As we ensemble two- and three-layered networks of corresponding domain (d) and input size (x), we observe a positive correlation between input size and accuracy. $\text{CNN}_{s,128}$ provides the maximum script identification rate of 91.41\% in this category. The primary reason behind the increase in accuracy is that the ensemble of two- and three-layered networks suggests that these networks complement each other. Further, to study an effect of the spatial (s) and frequency (f) domain representation in $\text{CNN}_{d,x,y}$, we ensemble networks across all input sizes and depth of network. The spatial representation, $\text{CNN}_{s}$, have produced script identification rate of 94.14\% and the frequency domain representation, $\text{CNN}_f$, have escalated up to 90.27\%. However, the frequency domain representations learning can be complimented and it has been clearly seen when we combined them. In their combination, we have achieved the highest script identification rate of 94.73\%. Since we have not received 100\% script identification rate, it is wise to provide a few samples where our system failed to identify correctly (see Fig.~\ref{fig:miss}). 

Like we have mentioned in Section~\ref{sec:data}, we also provide precision, recall and f-score for all architectures in Fig.~\ref{fig:result}. In what follows, the highest script identification rate, i.e. 94.73\% will be taken for a comparison. 

\subsubsection{Comparative study:} 
For a fair comparison, widely used deep learning methods, such as LeNet~\cite{lecun1998gradient} and AlexNet ~\cite{krizhevsky2012imagenet} were taken. In addition, recently reported work on 11 handwritten Indic script dataset~\cite{obaidullah2017phdindic_11} (including their baseline results) was considered. In Table~\ref{tab:comp}, we summarize the results. Our comparative study is focused on accuracy (not precision, recall and f-score), since other methods reported accuracy, i.e. identification rate. Of course, in Fig.~\ref{fig:result}, we are not just limited to accuracy. 

In Table~\ref{tab:comp}, our method outperforms all other methods. Precisely, it outperforms Obaidullah et al.~\cite{obaidullah2017phdindic_11} by 4.7\%, LeNet~\cite{lecun1998gradient} by 2.73\% and AlexNet~\cite{krizhevsky2012imagenet} by 2.61\%. 

\section{Conclusion}\label{sec:conc}
In this paper, we have proposed a novel framework that uses {\em c}onvolutional {\em n}eural {\em n}etworks (CNNs) for feature extraction. In our method, in addition, to conventional spatial domain representation, we have used multilevel 2D discrete Haar wavelet transform, where image representations have been scaled to a variety of different sizes. Having these, several different CNNs have been used to select features. With this, 11 different handwritten scripts: Bangla, Devnagari, Gujarati, Gurumukhi, Kannada, Malayalam, Oriya, Roman, Tamil, Telugu and Urdu, have been identified, where 1000 words per script are used. In our test, we have achieved the maximum script identification rate of 94.73\% using {\em m}ulti-{\em l}ayer {\em p}erceptron (MLP).  To the best of our knowledge, this is the biggest data for Indic script identification work. Considering the complexity and the size of the dataset, our method outperforms the previously reported techniques.

\begin{table}[tbp]
\centering
\renewcommand{\tabcolsep}{1.0em}
\renewcommand{\arraystretch}{1.2}
\caption{Comparative study.}
\label{tab:comp}
\begin{tabular}{lc}
\hline
\textbf{Method} & \textbf{Accuracy} \\
\hline \hline
Obaidullah et al.~\cite{obaidullah2017phdindic_11} & 91.00\% \\ 
(Hand-crafted features) & \\
LeNet ~\cite{lecun1998gradient} 	& 82.00\% \\
(CNN) & \\
AlexNet~\cite{krizhevsky2012imagenet} & 92.14\% \\
(CNN) & \\
Our method & 94.73\% \\
(Multiscale CNN + WT) \\ \hline
\end{tabular}
\end{table}

\bibliographystyle{splncs}
\bibliography{ref}

\begin{thebibliography}{10}

\bibitem{ghosh2010script}
Ghosh, D., Dube, T., Shivaprasad, A.:
\newblock Script recognition—a review.
\newblock IEEE Transactions on pattern analysis and machine intelligence
  \textbf{32}(12) (2010)  2142--2161

\bibitem{pal2012handwriting}
Pal, U., Jayadevan, R., Sharma, N.:
\newblock Handwriting recognition in indian regional scripts: a survey of
  offline techniques.
\newblock ACM Transactions on Asian Language Information Processing (TALIP)
  \textbf{11}(1) (2012) ~1

\bibitem{singh2015word}
Singh, P.K., Sarkar, R., Nasipuri, M., Doermann, D.:
\newblock Word-level script identification for handwritten indic scripts.
\newblock In: Document Analysis and Recognition (ICDAR), 2015 13th
  International Conference on, IEEE (2015)  1106--1110

\bibitem{hangarge2013directional}
Hangarge, M., Santosh, K., Pardeshi, R.:
\newblock Directional discrete cosine transform for handwritten script
  identification.
\newblock In: Document Analysis and Recognition (ICDAR), 2013 12th
  International Conference on, IEEE (2013)  344--348

\bibitem{PATI20081218}
Pati, P.B., Ramakrishnan, A.:
\newblock Word level multi-script identification.
\newblock Pattern Recognition Letters \textbf{29}(9) (2008)  1218 -- 1229

\bibitem{obaidullah2017phdindic_11}
Obaidullah, S.M., Halder, C., Santosh, K., Das, N., Roy, K.:
\newblock Phdindic\_11: page-level handwritten document image dataset of 11
  official indic scripts for script identification.
\newblock Multimedia Tools and Applications (2017)  1--36

\bibitem{lecun1998gradient}
LeCun, Y., Bottou, L., Bengio, Y., Haffner, P.:
\newblock Gradient-based learning applied to document recognition.
\newblock Proceedings of the IEEE \textbf{86}(11) (1998)  2278--2324

\bibitem{roy2017handwritten}
Roy, S., Das, N., Kundu, M., Nasipuri, M.:
\newblock Handwritten isolated bangla compound character recognition: A new
  benchmark using a novel deep learning approach.
\newblock Pattern Recognition Letters \textbf{90} (2017)  15--21

\bibitem{krizhevsky2012imagenet}
Krizhevsky, A., Sutskever, I., Hinton, G.E.:
\newblock Imagenet classification with deep convolutional neural networks.
\newblock In: Advances in neural information processing systems. (2012)
  1097--1105

\bibitem{sarkhel2017multi}
Sarkhel, R., Das, N., Das, A., Kundu, M., Nasipuri, M.:
\newblock A multi-scale deep quad tree based feature extraction method for the
  recognition of isolated handwritten characters of popular indic scripts.
\newblock Pattern Recognition (2017)

\bibitem{smith1997scientist}
Smith, S.W.,  et~al.:
\newblock The scientist and engineer's guide to digital signal processing.
\newblock (1997)

\bibitem{daubechies1990wavelet}
Daubechies, I.:
\newblock The wavelet transform, time-frequency localization and signal
  analysis.
\newblock IEEE transactions on information theory \textbf{36}(5) (1990)
  961--1005

\bibitem{portnoff1980time}
Portnoff, M.:
\newblock Time-frequency representation of digital signals and systems based on
  short-time fourier analysis.
\newblock IEEE Transactions on Acoustics, Speech, and Signal Processing
  \textbf{28}(1) (1980)  55--69

\bibitem{sundararajan2011fundamentals}
Sundararajan, D.:
\newblock Fundamentals of the discrete haar wavelet transform.
\newblock (2011)

\bibitem{vonesch2007generalized}
Vonesch, C., Blu, T., Unser, M.:
\newblock Generalized daubechies wavelet families.
\newblock IEEE Transactions on Signal Processing \textbf{55}(9) (2007)
  4415--4429

\bibitem{kingma2014adam}
Kingma, D., Ba, J.:
\newblock Adam: A method for stochastic optimization.
\newblock arXiv preprint arXiv:1412.6980 (2014)

\end{thebibliography}

\end{document}